\def\year{2019}\relax
\documentclass[letterpaper]{article} %DO NOT CHANGE THIS
\usepackage{aaai19}  %Required
\usepackage{times}  %Required
\usepackage{helvet}  %Required
\usepackage{courier}  %Required
\usepackage{url}  %Required
\usepackage{graphicx}  %Required
\usepackage{enumerate}
\usepackage[linesnumbered,ruled]{algorithm2e}
\usepackage{array}
\usepackage{subfig}
\usepackage{multirow}
\usepackage{ctable}
\begin{document}
\setcounter{secnumdepth}{2}
\title{Automatically Evolving CNN Architectures Based on Blocks}
\author{Yanan Sun \and Bing Xue \and Mengjie Zhang \and Gary G. Yen
}
\maketitle
\begin{abstract}
The performance of Convolutional Neural Networks (CNNs) highly relies on their architectures. In order to design a CNN with promising performance, extended expertise in both CNNs and the investigated problem is required, which is not necessarily held by every user interested in CNNs or the problem domain. In this paper, we propose to automatically evolve CNN architectures by using a genetic algorithm based on ResNet blocks and DenseNet blocks. The proposed algorithm is \textbf{completely} automatic in designing CNN architectures, particularly, neither pre-processing before it starts nor post-processing on the designed CNN is needed. Furthermore, the proposed algorithm does not require users with domain knowledge on CNNs, the investigated problem or even genetic algorithms. The proposed algorithm is evaluated on CIFAR10 and CIFAR100 against 18 state-of-the-art peer competitors. Experimental results show that it outperforms state-of-the-art CNNs hand-crafted and CNNs designed by automatic peer competitors in terms of the classification accuracy, and achieves the competitive classification accuracy against semi-automatic peer competitors. In addition, the proposed algorithm consumes much less time than most peer competitors in finding the best CNN architectures. 
\end{abstract}
\section{Introduction}
\noindent Convolutional Neural Networks (CNNs) have demonstrated their promising performance on various real-world applications~\cite{sainath2013deep,sutskever2014sequence,alphago}. However, it has been known that the performance of CNNs highly depends on their architectures. Mathematically, a CNN can be formulated as Equation~(\ref{equ_cnn})
\begin{equation}
\label{equ_cnn}
\left\{
\begin{array}{rl}
	a =& \mathcal{A}(x, y)\\
	w =& \mathcal{G}(a)\\
	z = & \mathcal{F}(w, x, y)
\end{array}
\right.
\end{equation}
where $a$ is the architecture designed by the method $\mathcal{A}(\cdot)$ for the data $(x,y)$; $w$ denotes the weights of the CNN based on the architecture as well as the used weight initialization method $\mathcal{G}(\cdot)$; $z$ refers to the features learned by the CNN based on the weights and the data, and $\mathcal{F}(\cdot)$ is the composition of the sequential operations in the CNN, such as the convolutional operation, pooling operation, non-linear activation functions, and to name a few. 

In the case of a classification task to which the CNN is applied, a classifier is added to the tail of the CNN for receiving the learned features $z$. In order to let $z$ be more meaningful, i.e., $z$ could significantly contribute to the classification performance, the weights $w$ is optimized based on the objective function of the used classifier. Typically, the objective function is differentiable, resulting in the gradient-based approaches being the dominant weight optimization algorithm. However, the architecture design method $\mathcal{A}(\cdot)$ is non-convex or non-differentiable, and accurate methods are \textbf{incapable} of finding the best architecture for the given data. To this end, state-of-the-art CNNs are still hand-crafted, e.g., ResNet~\cite{he2016deep} and DenseNet~\cite{huang2017densely}. Unfortunately, hand-crafting CNNs requires the expertise in both the investigated data and CNNs, which is not necessarily held by the users who want to experience the powerful functions of CNNs. Consequently, there is a high demand in algorithms being able to automatically design\footnote{In this paper, the terms ``design'', ``find'', ``learn'' and ``evolve'' are with the same meaning when they are used together with ``CNN architectures''.} CNN architectures without such expertise.

Algorithms specially developed for designing CNN architectures are mainly proposed in recently two years. Based on whether pro- or post-processing is required in using these algorithms for designing CNNs, they can be divided into two different categories, the semi-automatic CNN architecture design algorithms and the completely automatic ones. Particularly, the semi-automatic algorithms cover the genetic CNN method (Genetic CNN)~\cite{xie2017genetic}, the hierarchical representation-based method (Hierarchical Evolution)~\cite{liu2017hierarchical}, the efficient architecture search method (EAS)~\cite{cai2018efficient}, and the block design method (Block-QNN-S)~\cite{zhong2017practical}. The automatic algorithms include the large-scale evolution method (Large-scale Evolution)~\cite{real2017large}, the Cartesian genetic programming method (CGP-CNN)~\cite{suganuma2017genetic}, the neural architecture search method (NAS)~\cite{zoph2016neural}, and the meta-modelling method (MetaQNN)~\cite{baker2016designing}. These algorithms are mainly based on evolutionary algorithms~\cite{back1996evolutionary} or reinforcement learning~\cite{sutton1998reinforcement}. Specifically, Genetic CNN, Large-scale Evolution, Hierarchical Evolution and CGP-CNN are based on evolutionary algorithms, and NSA, MetaQNN, EAS and Block-QNN-S are based on reinforcement learning.

Experimental results from these algorithms have shown their promising performance in finding the best CNN architectures. However, major limitations still exist. Firstly, expertise in the investigated data and CNNs is still needed by the semi-automatic CNN architecture design algorithms. For example, EAS takes effect on a base network which already has fairly good performance on the investigated problem. However, the base network is manually designed based on expertise. Block-QNN-S only designs several small networks, and these networks are then integrated into a larger CNN framework. However, the other types of layers, such as the pooling layers, need to be assigned into the CNN framework with expertise. Secondly, the CNN architecture design algorithms based on reinforcement learning typically consume much more computational resource. For instance, NAS consumes 28 days on 800 Graphic Process Unit (GPU) cards for the CIFAR10 dataset~\cite{krizhevsky2009learning}. However, sufficient computation resource is not necessarily available to every interested user. Thirdly, the CNN architecture design algorithms based on evolutionary algorithms use only partial functions of the evolutionary algorithms, which results in the CNNs which are usually not with the promising performance for the investigated problems. For example, Genetic CNN employs a fixed-length encoding scheme to represent CNNs. However, we never know the best depth of the CNN in solving a new problem. To this end, Large-scale Evolution utilizes a variable-length encoding scheme where the CNNs can adaptively change their depths for the problems. However, Large-scale Evolution uses only the mutation operator but not anys crossover operator during the search process. In evolutionary algorithms, the crossover operator and mutation operator play complementary roles of local search and global search. The best performance of evolutionary algorithms needs to use both operators. It is not strange that  Large-scale Evolution does not use the crossover operator since the crossover operator is originally designed for the fixed-length encoding scheme.

To this end, the development of CNN architecture design algorithms, especially for the automatic ones with promising performance but relying on the limited computational resource, is still in their infancy. The aim of this paper is to design and develop a new genetic method-based algorithm to automatically design CNN architectures by addressing the limitations discussed above. To achieve this goal, the objectives below have been specified:
\begin{itemize}
	\item Neither base CNN nor expertise in the investigated data and CNNs is required before using the proposed algorithm. The CNN whose architecture is designed by the proposed algorithm can be directly used without any re-composition, pre-processing, or post-processing.
	
	\item The variable-length encoding scheme is employed for the unpredictably optimal depth of the CNN, and a new crossover operator and a mutation operator are both incorporated into the proposed algorithm to jointly exploit and explore the best CNN architectures.
	
	\item An efficient encoding strategy is designed based on the ResNet and DenseNet blocks for speeding up the architecture design, and limited computational resource is utilized, while the promising performance is still achieved by the proposed algorithm.
\end{itemize}

The remainder of the paper is organized as follows. The background related to the proposed algorithm is introduced in Section~\ref{section2}. Then, the details of the proposed algorithm are documented in Section~\ref{section3}. To evaluate the performance of the proposed algorithm, the experiment design and the experimental results are shown in Sections~\ref{section4} and \ref{section5}, respectively. Finally, the conclusions and future work are summarized in Section~\ref{section6}.

\section{Background}
\label{section2}
In this section, Genetic Algorithms (GAs), ResNet Blocks (RBs) and DenseNet Blocks (DBs), which are the base work of the proposed algorithm, are introduced to help readers more easily understand the details of the proposed algorithm.

\subsection{Genetic Algorithms}
GAs~\cite{holland1992adaptation} are a class of heuristic population-based paradigm. They are also the most popular type of evolutionary algorithms (evolutionary algorithms also include genetic programming~\cite{banzhaf1998genetic}, evolutionary strategy~\cite{janis1976evolutionary} and so on, in addition to GAs). Because of the nature of gradient-free, GAs are preferred especially in engineering fields where the optimization problems are commonly non-convex and non-differentiable~\cite{deb2002fast,sun2018igd,sun2018improved,sun2017reference}. GAs address optimization problems by imitating the biological evolution through a series of bio-inspired operators, such as crossover, mutation and selection. Commonly, a GA works as follows:
\begin{enumerate}[\textit{Step} 1:]
	\item Initialize a population of individuals which represent a number of solutions of the problem through the employed encoding strategy; \label{ga_step1}
	\item Evaluate the fitness of each individual on the encoded solution in the population; \label{ga_step2}
	\item Select promising parent individuals from the current population, and then generate offspring with crossover and mutation operators; \label{ga_step3}
	\item Evaluate the fitness of the generated offspring; \label{ga_step4}
	\item Select a population of individuals with promising performance from the current population, and then remove the current population by the selected population; \label{ga_step5}
	\item Go to Step~\ref{ga_step3} if the evolution is not terminated; otherwise select the individual with the best fitness as the best solution for the problem.
\end{enumerate}
Commonly, a maximal generation number is predefined as the termination criterion.
\subsection{ResNet and DenseNet Blocks}
ResNet~\cite{he2016deep} and DenseNet~\cite{huang2017densely} are two state-of-the-art CNNs. The success of ResNet and DenseNet largely owns to their building blocks, i.e., RBs and DBs, respectively. 
\begin{figure}
	\centering
	\includegraphics[width=\columnwidth]{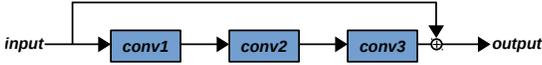}
	\caption{An example of the ResNet block (RB).}
	\label{fig_resnet_block}
\end{figure}

\begin{figure}
	\centering
	\includegraphics[width=\columnwidth]{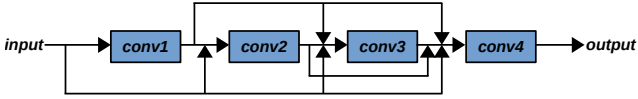}
	\caption{An example of the DenseNet block (DB) including four convolutional layers.}
	\label{fig_dense_block}
\end{figure}

Fig.~\ref{fig_resnet_block} shows an example of an RB which is composed of three convolutional layers\footnote{Here we only detail this type of block which is used to build deeper networks. Indeed, ResNet also has another type of block which is typically used for building networks with no more than 34 layers.} and one skip connection. In this example, the convolutional layers are denoted as $conv1$, $conv2$ and $conv3$. On $conv1$, the spatial size of the input is reduced by a smaller number of filters with the size of $1\times 1$, to lower the computational complexity of $conv2$. On $conv2$, filters with a larger size, such as $3\times 3$, are used to learn features with the same spatial size. On $conv3$, filters with the size of $1\times1$ are used again, and the spatial size is increased for generating more features. The input is added, denoted by $\oplus$, to the output of $conv3$ as the final output of the RD. Note that, if the spatial sizes of input and $conv3$'s output are unequal, a group of convolutional operation with the filters of $1\times1$ size is used for the input, to achieve the same spatial size as that of $conv3$'s output, for the addition.

Fig.~\ref{fig_dense_block} exhibits an example of a DB. For the convenience of the introduction, we give only four convolutional layers in the DB. In practise, a DB can have a different number of convolutional layers, which is tuned by users. In a DB, each convolutional layer receives inputs from not only the input data but also the output of all the previous convolutional layers. In addition, there is a parameter, $k$, for controlling the spatial size of the input and output of the same convolutional layer. If the spatial size of the input is $a$, then the spatial size of the output is $a+k$, which is achieved by the convolutional operation using the corresponding number of filters.

Researchers have tried to unveil the principles of RBs and DBs behind their success. However, a completely satisfactory explanation is still elusive~\cite{orhan2017skip}. Because our goal in this paper is to automatically find the best CNN architectures based on RBs and DBs, we will not put details on attempting the principles.

\section{The Proposed Algorithm}
\label{section3}
In this section, the framework of the proposed algorithm and its main components are detailed. For the convenience of the development, the proposed algorithm is named AE-CNN (Automatically Evolving CNNs) in short, and the evolved CNN is used for image classification tasks.
\subsection{Algorithm Overview}
\begin{algorithm}
	\caption{Framework of AE-CNN}\label{alg_framework}
	\KwIn{The population size $N$, the maximal generation number $T$, the crossover propability $\mu$, the mutation probability $\nu$.}
	\KwOut{The best CNN.}
	$P_0 \leftarrow$Initialize a population with the size of $N$ by using the proposed encoding strategy;\\
	$t\leftarrow 0$;\\
	\While{$t < T$}
	{
		Evaluate the fitness of individuals in $P_t$;\\
		\label{alg_framework_fitness1}
		$Q_t\leftarrow\emptyset$;\\
		\While{$|Q_t| < N$}
		{
			$p_1,p_2\leftarrow$ Select two parent individuals from $P_t$ by using binary tournament selection;\\
			\label{alg_framework_parent_selection}
			$q_1, q_2\leftarrow$ Generate two offspring by $p_1$ and $p_2$ by crossover operation with the probability of $\mu$ and mutation operation with the probability of $\nu$;\\
			$Q_t\leftarrow Q_t\cup q_1\cup q_2$;\\
		}
		Evaluate the fitness of individuals in $Q_t$;\\
		\label{alg_framework_fitness2}
		$P_{t+1}\leftarrow$ Select $N$ individuals from $P_t\cup Q_t$ by environmental selection;\\
		\label{alg_framework_env_selection}
		$t\leftarrow t+1$;
	}
	Select the best individual from $P_t$ and decode it to the corresponding CNN.
	
\end{algorithm}

Algorithm~\ref{alg_framework} shows the framework of AE-CNN. Because AE-CNN follows the standard process of GAs as discussed in Section~\ref{section2}, we only briefly document the fitness evaluation (lines~\ref{alg_framework_fitness1} and~\ref{alg_framework_fitness2}), selection of parent individuals (line~\ref{alg_framework_parent_selection}) and environmental selection (line~\ref{alg_framework_env_selection}), which are the same as GAs; while the new encoding strategy, the new crossover operation and the new mutation operation, which are the main difference of AE-CNN to existing peer competitors, are detailed in Subsections~\ref{section3_encode}, \ref{section3_crossover} and~\ref{section3_mutation}, respectively.

The fitness of the individuals provides a quantitative measurement indicating how well they adapt to the environment, and is calculated based on the information these individuals encoded and the task to be solved. In AE-CNN, the fitness of an individual is the classification accuracy based on the architecture encoded by the individual and the corresponding ``validation'' dataset. According to the principle of evolutionary algorithms, an individual with a high fitness has a high probability to generate an offspring hopefully with even higher fitness than itself. Each individual in AE-CNN is decoded to the corresponding CNN, and then added to a classifier to be trained like that of a common CNN. Typically, the widely used classifier is the Logistic regression for binary classification and the Softmax regression for multiple classification. In AE-CNN, the decoded CNN is trained on the ``training'' dataset, and the fitness is the best classification accuracy on the ``validation'' dataset during the CNN training.

In order to generate a population of offspring, parent individuals need to be chosen in advance. In AE-CNN, the binary tournament selection~\cite{miller1995genetic} is used for this purpose, based on the conventions of the GA community. The binary tournament selection randomly selects two individuals from the population, and the one with a higher fitness is chosen. By performing $N/2$ times of the selection, $N$ individuals are selected, and then $N$ offspring are generated. 

In the environmental selection, a population of individuals with the size of $N$ is selected from the current population, i.e., $P_t\cup Q_t$, serves as the parent individuals for the next generation. Theoretically, a good population has the characteristics of both convergence and diversity~\cite{holland1992adaptation}, to prevent from trapping into local minima and premature convergence. In practice, the parent individuals should be composed of individuals with the best fitness for the convergence, and individuals whose fitness have significant differences from each other for the diversity. To this end, we will purposely select the individual with the best fitness, along with $N-1$ individuals which are selected by binary tournament selection, to generate offspring for the new population as parent individuals.

\subsection{Encoding Strategy}
\label{section3_encode}
The proposed encoding strategy aims at effectively modelling CNNs with different architectures by individuals in the used GA. Typically, the architecture of a CNN is decided by multiple convolutional layers, pooling layers and fully-connected layers with a particular order, as well as their parameter settings. In the proposed algorithm, CNNs are constructed based on RBs, DBs and pooling layers, which is motivated by the remarkable success of ResNet and DenseNet. The fully-connected layers are not considered in this proposed algorithm. The main reason is that the fully-connected layers easily cause the over-fitting phenomenon due to their full-connection nature. To control this phenomenon, other techniques must be incorporated, such as Dropout~\cite{srivastava2014dropout}. However, these techniques will also give rise to extra parameters that need to be carefully tuned, which will increase the computational complexity of the proposed algorithm. The experimental results shown in Section~\ref{section5} will justify that the promising performance can be still achieved without using the fully-connected layers.

For the used RBs, based on the configuration of state-of-the-art CNNs~\cite{szegedy2015going,he2016deep}, we set the filter size of $conv2$ to $3\times 3$, which is also used for the convolutional layers in the used DBs. For the used pooling layers, we set the same stride as the step size to $2\times 2$ based on the conventions, which means that such a single pooling layer in the evolved CNN halves the input dimension for one time. To this end, the unknown parameter settings for RBs are the spatial sizes of input and output, those for DBs are the spatial sizes of input and output, as well as $k$, and that for pooling layers are only their types, i.e., the $max$ or $mean$ pooling type. Note that the number of convolutional layers in a DB is known because it can be derived by the spatial sizes of input and output as well as $k$.

Accordingly, the proposed encoding strategy is based on three different types of units and their positions in the CNNs. The units are the RB Unit (RBU), the DB Unit (DBU) and the Pooling layer Unit (PU). Specifically, an RBU and a DBU contain multiple RBs and DBs, respectively, while a PU is composed of only a single pooling layer. Our motivation is that: 1) by putting multiple of RBs or DBs into an RBU or a DBU, the depth of the CNN can be significantly changed compared to stacking RBs or DBs one by one, which will speed up the heuristic search of the proposed algorithm by easily changing the depth of the CNN; and 2) one PU carrying a single pooling layer is more flexible than carrying multiple pooling layers, because the effect of multiple consequent pooling layers can be achieved by stacking multiple PUs. In addition, we also add one parameter to represent the unit type. In summary, the encoded information for an RBU are the type, the number of RBs, the input spatial size and the output spatial size, which are denoted as $type$, $amount$, $in$ and $out$, respectively. The encoded information of a DBU is the same as those of an RBU, in addition to the additional parameter $k$. Only one parameter is needed in a PU for encoding the pooling type.   

\begin{figure}[htp]
	\centering
	\includegraphics[width=\columnwidth]{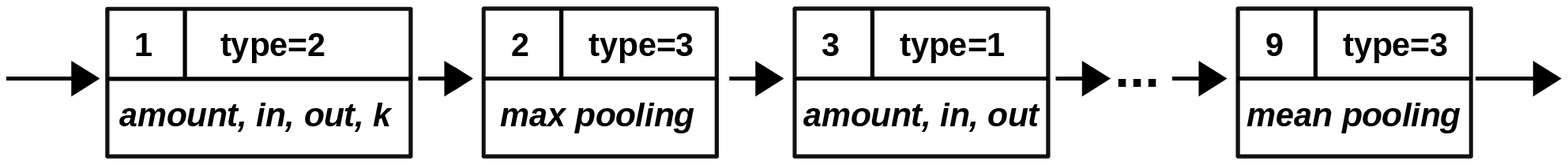}
	\caption{An example of the proposed encoding strategy.}
	\label{fig_encoding}
\end{figure}

Fig.~\ref{fig_encoding} shows an example of the proposed algorithm in encoding a CNN containing nine units. Specifically, each number in the upper-left corner of the block denotes the position of the unit in the CNN. The unit is an RUB, a DBU or a PU if the $type$ is $1$, $2$, or $3$, respectively. Note that we do not constrain the maximal length of each individual, which means that the proposed algorithm can adaptively find the best CNN architecture with a proper depth through the designed variable-length encoding strategy.

\subsection{Crossover Operation}
\label{section3_crossover}
In traditional GAs, the crossover operation is performed on two individuals with the same length, which is biologically evidenced. Based on the proposed encoding strategy, individuals in the proposed algorithm have different lengths, i.e., the corresponding CNNs are with different depths. In this regard, the traditional crossover operator cannot be used. However, the crossover operator often refers to the local search ability of GAs, exploiting the search space for a promising performance. The performance of the final solution may be deteriorated due to the lacking of the crossover operation in GAs. 

\begin{algorithm}
	\caption{Crossover Operation of AE-CNN}\label{alg_crossover}
	\KwIn{Two parent individuals, $p_1$ and $p_2$, selected by the binary tournament selection, crossover propability $\mu$.}
	\KwOut{Two offspring.}
	$r\leftarrow$ Uniformly generate a number from $[0, 1]$;\\
	\uIf{$r < \mu$}
	{
		Randomly choose a position from $p_1$ and $p_2$;\\
		Separate $p_1$ and $p_2$ based on the chosen positions;\\
		$q_1\leftarrow$ Combine the first part of $p_1$ and the second part of $p_2$;\\
		$q_2\leftarrow$ Combine the first part of $p_2$ and the first part of $p_1$;\\
	}
	\Else
	{
		$q_1\leftarrow p_1$;\\
		$q_2\leftarrow p_2$;\\
	}
	\textbf{Return} $q_1$ and $q_2$.
\end{algorithm}

In the proposed algorithm, we employ the one-point crossover operator. The reason is that the one-point crossover has been widely used in Genetic Programming (GP)~\cite{banzhaf1998genetic}. GP is another important type of evolutionary algorithms, and the individuals in GP are commonly with different lengths. Algorithm~\ref{alg_crossover} shows the crossover operation in the proposed algorithm.

Note that some changes are automatically performed on the generated offspring if required. For example, the $in$ of the current unit should be equal to the $out$ of the previous unit, and other cascade adjustments caused by this change. 

\subsection{Mutation Operation}
\label{section3_mutation}
The mutation operation typically performs the global search in GAs, exploring the search space for promising performance. It works on each generated offspring with a predefined probability and the allowed mutation types. Available mutation types are designed based on the proposed encoding strategy. In the proposed algorithm, the available mutation types are: 
\begin{itemize}
	\item Adding (adding an RBU, adding a DBU, or adding a PU to the selected position);
	\item Removing (removing the unit at the selected position);
	\item Modifying (modifying the encoded information of the unit at the selected position).
\end{itemize}

\begin{algorithm}
	\caption{Mutation Operation of AE-CNN}\label{alg_mutation}
	\KwIn{The offspring $q_1$, mutation propability $\nu$.}
	\KwOut{The mutated offspring.}
	$r\leftarrow$ Uniformly generate a number from $[0, 1]$;\\
	\uIf{$r < \nu$}
	{
		Randomly choose a position from $q_1$;\\
		$type\leftarrow$ Randomly select one from \{Adding, Removing, Modifying\};\\
		\uIf{$type$ is Adding}
		{
			$mu\leftarrow$ Randomly select one from \{adding an RBU, adding a DBU, adding a PU\}
		}
		\uElseIf {$type$ is Removing}
		{
			$mu\leftarrow$ removing a unit;
		}
		\Else
		{
			$mu\leftarrow$ modifying the encoded information;
		}
		Perform $mu$ at the chosen position;\\
	}
	\textbf{Return} $q_1$.
\end{algorithm}

The mutation operation in the proposed algorithm is detailed in Algorithm~\ref{alg_mutation}. Because all the generated offspring use the same routine for the mutation, Algorithm~\ref{alg_mutation} shows only the process of one offspring for the reason of simplicity. Note that the offspring will be kept the same if it is not mutated. In addition, a series of necessary adjustments will also be automatically performed based on the logic of composing a valid CNN as highlighted in the crossover operation.

\section{Experiment Design}
\label{section4}
\subsection{Peer Competitors}
In order to show the superiority of the proposed algorithm, various peer competitors are chosen to perform the comparison. Particularly, the chosen peer competitors can be divided into three different categories. 

The first includes the state-of-the-art CNNs whose architectures are manually designed with extensive expertise: DenseNet~\cite{huang2017densely}, ResNet~\cite{he2016deep},  Maxout~\cite{goodfellow2013maxout}, VGG~\cite{simonyan2014very}, Network in Network~\cite{lin2013network}, Highway Network~\cite{srivastava2015highway}, All-CNN~\cite{springenberg2014striving} and FractalNet~\cite{larsson2016fractalnet}. In addition, considering the promising performance of ResNet, we use two versions of ResNet in the experiment, they are the ResNet with 101 layers and ResNet with 1,202 layers, which are denoted as ResNet (depth=101) and ResNet (depth=1,202), respectively. 

The second covers the CNN architecture design algorithms with a semi-automatic way, including Genetic CNN~\cite{xie2017genetic}, Hierarchical Evolution~\cite{liu2017hierarchical}, EAS~\cite{zoph2016neural}, and Block-QNN-S~\cite{zhong2017practical}.

The third refers to Large-scale Evolution~\cite{real2017large}, CGP-CNN~\cite{suganuma2017genetic}, NAS~\cite{zoph2016neural}, and MetaQNN~\cite{baker2016designing}, which design CNN architectures in a completely automatic way.

\begin{figure}[!htp]
	\centering
	\subfloat[CIFAR10]{\includegraphics[width=0.85\columnwidth]{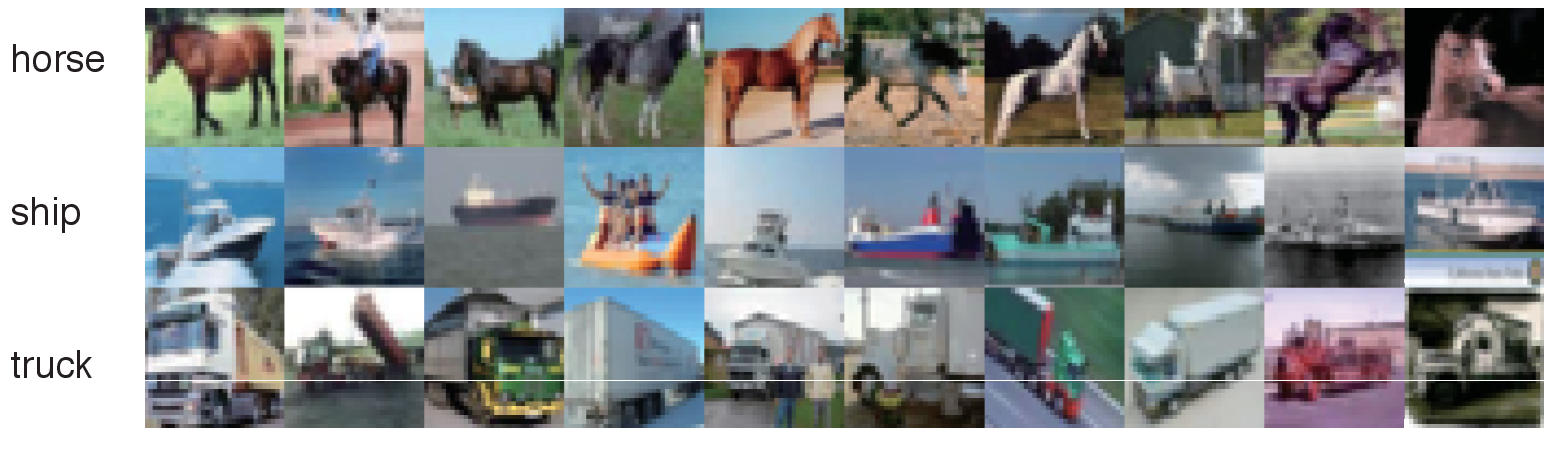}\label{fig_cifar10_example}}
	\hfil
	\subfloat[CIFAR100]{\includegraphics[width=0.85\columnwidth]{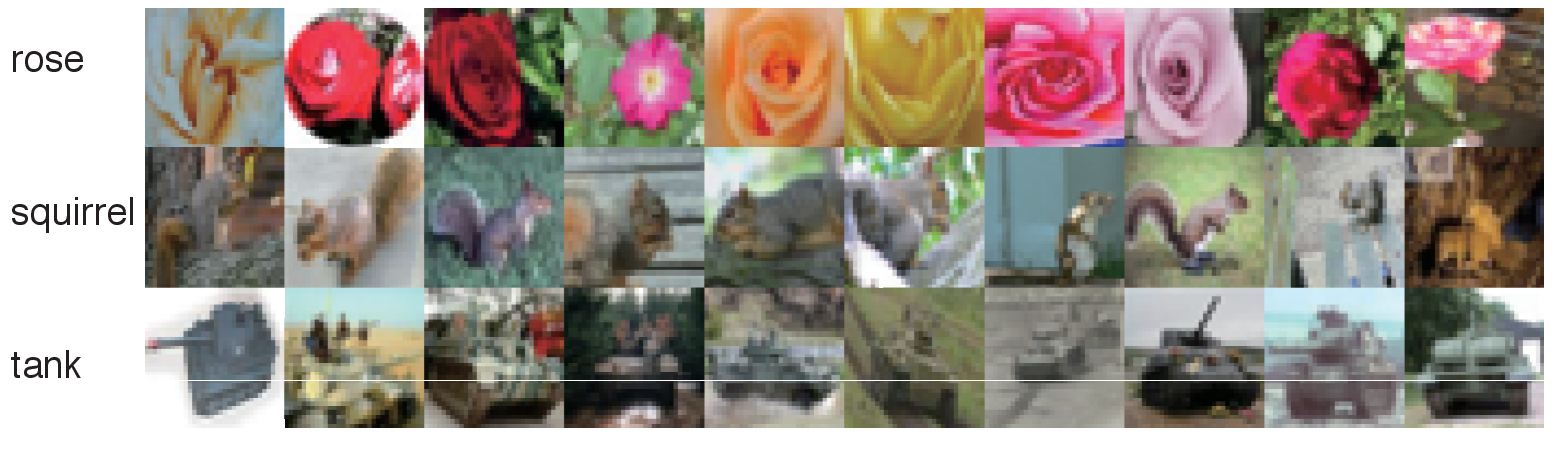}\label{fig_cifar100_example}}
	\caption{Examples from CIFAR10 (shown in Fig.~\ref{fig_cifar10_example}) and CIFAR100 (shown in Fig.~\ref{fig_cifar100_example}).}\label{fig_benchmark_example}
\end{figure}

\subsection{Benchmark Datasets}
CNN architecture design algorithms typically perform image classification tasks by using their designed CNNs to compare the classification performance, where CIFAR10 and CIFAR100~\cite{krizhevsky2009learning} are the most popular benchmark datasets. Therefore, both datasets are also used in this experiment for quantifying the performance of the proposed algorithm.

CIFAR10 and CIFAR100 are two widely used image classification benchmark datasets for recognizing nature objects, such as bird, boat and air plane. Each set has $50,000$ training images and $10,000$ test images. The differences between CIFAR10 and CIFAR100 are that CIFAR10 is for 10 classes classification while CIFAR100 is for 100 classes. However, each benchmark has nearly the same number of training images for each class. 

Fig.~\ref{fig_benchmark_example} illustrates the images from each benchmark for reference, where images in each row denote the ones from the same class, and the words in the left column refer to the corresponding class name. In addition, the training dataset is augmented by padding four zeros to each side of one image, and then randomly cropped to the original size followed by a randomly horizontal flip, based on the conventions of the peer competitors.

\subsection{Parameter Settings}
In the comparison, we use the results of the peer competitors reported in their seminal papers rather than performing them by us. The reason is that the results reported by the them are usually the best. To this end, there is no need to set the parameters of the peer competitors. For the proposed algorithm, the population size and maximal generation number are set to be $20$, the probabilities of crossover and mutation are set to $0.9$ and $0.2$, respectively, based on the conventions of the GA community. The validation dataset is randomly split from the training dataset with the proportion of 1/5 based on the convention of machine learning community. All algorithms are evaluated on the same test dataset.

Each individual is trained by Stochastic Gradient Descend (SGD) with a batch size of 128. The parameter settings for SGD are also based on the conventions from peer competitors. Specifically, the momentum is set to 0.9. The learning rate is initialized to $0.01$, but with a warming setting of $0.1$ during the second to the $150$-th epoch, and scaled by diving 10 at the $250$-th epoch. The weight decay is set to $5\times10^{-4}$. In addition, the fitness of the individual is set to zero if it is out of memory during the training. When the evolutionary process terminates, the best individual is retrained on the original training dataset with the same SGD settings, and the test accuracy is reported on the test dataset for the comparison. Considering the heuristic nature of the proposed algorithm as well as the expensive computational cost, the best individual is trained for five independent runs. In the deep learning community, the convention is to report only the best result. Therefore, the comparison result of the proposed algorithm is also the best result from the five results.

In addition, the available choices of $k$ in a DB are $12$, $20$ and $40$ based on the design of DenseNet, the maximal convolutional layers in a DB are specified as $10$ (when $k=12$ and $k=20$) and $5$ (when $k=40$). Both the maximal numbers of RBU and DBU in a CNN are set to $6$. Note that these settings are mainly based on our available computational resources because any number beyond these settings will be easily out of the memory. 
All the experiments are performed on two GPU cards with the model of Nvidia GeForce GTX 1080 Ti.

\section{Experimental Results}
\label{section5}

\begin{table*}
	\caption{The comparisons between the proposed algorithm and the state-of-the-art peer competitors in terms of the classification error (\%), number of parameters and the consumed GPU Days on the CIFAR10 and CIFAR100 benchmark datasets.}
	\label{tab_overview}
	\center
	\begin{tabular}{lcccc}
		\hline
		& \textbf{CIFAR10} & \textbf{CIFAR100} & \textbf{\# of Parameter} & \textbf{GPU Days} \\
		\hline
		DenseNet (k=12)~\cite{huang2017densely} & 5.24& 24.42 & 1.0M & -- \\
		ResNet (depth=101)~\cite{he2016deep} &  6.43 & 25.16 & 1.7M & -- \\
		ResNet (depth=1,202)~\cite{he2016deep} & 7.93 & 27.82 & 10.2M & --\\
		Maxout~\cite{goodfellow2013maxout} & 9.3 & 38.6 & -- & -- \\
		VGG~\cite{simonyan2014very} & 6.66 & 28.05 & 20.04M & -- \\
		Network in Network~\cite{lin2013network} & 8.81 & 35.68 & -- & -- \\
		Highway Network~\cite{srivastava2015highway} & 7.72 & 32.39 & -- & -- \\
		All-CNN~\cite{springenberg2014striving} & 7.25 & 33.71 & -- & -- \\
		FractalNet~\cite{larsson2016fractalnet} & 5.22 & 22.3 & 38.6M & -- \\
		\hline
		Genetic CNN~\cite{xie2017genetic} & 7.1 & 29.05 &-- & 17 \\
		Hierarchical Evolution~\cite{liu2017hierarchical} & 3.63 & -- &  -- & 300 \\
		EAS~\cite{zoph2016neural} & 4.23 & -- & 23.4M & 10 \\
		Block-QNN-S~\cite{zhong2017practical} & 4.38 & 20.65 & 6.1M & 90 \\
		\hline
		Large-scale Evolution~\cite{real2017large} & 5.4 & -- & 5.4M &2,750 \\
		Large-scale Evolution~\cite{real2017large} & -- & 23 & 40.4M &2,750 \\
		CGP-CNN~\cite{suganuma2017genetic} & 5.98 & -- & 2.64M & 27 \\
		NAS~\cite{zoph2016neural} & 6.01 & -- & 2.5M & 22,400 \\
		MetaQNN~\cite{baker2016designing} & 6.92 & 27.14 & -- & 100 \\
		\hline
		AE-CNN & 4.7 & -- & 10.4M & 66 \\
		AE-CNN & -- & 22.4 & 8M & 38 \\
		\hline
		
	\end{tabular}
\end{table*}
In the experiments, we investigate not only the classification error, but also the number of parameters as well as the computational complexity for a comprehensive comparison. Because it is hard to theoretically analyze the computational complexity of each peer competitor, the consumed ``GPU Days'' is used as an indicator of the computational complexity. Specifically, the number of GPU Days is calculated by multiplying the number of employed GPU cards and the days the algorithms performed for finding the best architectures. For example, the proposed algorithm totally performed 33 days on two GPU cards for the CIFAR10 dataset, therefore, its GPU Days is $66$ by multiplying $33$ by $2$. 

Table~\ref{tab_overview} shows the experimental results of the propose algorithm as well as the peer competitors. In order to conveniently investigate the comparisons, Table~\ref{tab_overview} is divided into five ``rows'' by six horizontal lines. The first denotes the title of each column, the second, third and fourth rows refer to the state-of-the-art peer competitor whose architectures are manually designed, the semi-automatic and automatic CNN architecture design peer competitors, respectively. The fifth row shows the results of the proposed algorithm which is an automatic algorithm in designing CNN architectures. In addition, the symbol ``--'' in Table~\ref{tab_overview} implies there is no result publicly reported by the corresponding peer competitor.

As shown in Table~\ref{tab_overview}, AE-CNN outperforms all the state-of-the-art peer competitors manually designed on CIFAR10. Specifically, AE-CNN achieves the classification error of around 0.5\% lower than DenseNet (k=12) and FractalNet, $2.0\%$ lower than ResNet (depth=101), VGG and All-CNN, $3.0\%$ lower than ResNet (depth=1,202) and Highway Network, and even $4.5\%$ lower than Maxout and Network in Network. On CIFAR100, AE-CNN shows significantly lower classification error than Maxout, Network in Network, Highway Network and All-CNN, slightly better classification error than DenseNet (k=12), ResNet (depth=101), ResNet (depth=1,202) and VGG, and similar performance to FractalNet. The number of parameters of the CNN evolved by AE-CNN on both CIFAR10 and CIFAR100 is larger than DenseNet (k=12) and ResNet (depth=101), but nearly the same to ResNet (depth=1,202) and much less than that of VGG and FractalNet.

Among the semi-automatic peer competitors, AE-CNN performs better than Genetic CNN on both CIFAR10 and CIFAR100. Although Hierarchical Evolution and Block-QNN-S show better performance than AE-CNN on CIFAR10 and CIFAR100, AE-CNN consumed 1/5 GPU days of that consumed by Hierarchical Evolution on CIFAR10, and 1/3 GPU Days of that consumed by Block-QNN-S on CIFAR100. In addition, EAS and AE-CNN perform nearly the same on CIFAR10, while the best CNN evolved by AE-CNN only has 10.4M parameters, which is the half of that from EAS. In summary, AE-CNN shows the competitive performance among the semi-automatic peer competitors. It is important to note that the extra expertise is still required when using the algorithms from this category. For example, EAS only consumes 10 GPU Days for the best CNN on CIFAR10, which is based on a base CNN which already is with fairly good performance. So the comparison is not entirely fair to the proposed AE-CNN algorithm, which is completely automatic without using any human expertise and/or extra resources. 

Among the automatic peer competitors, AE-CNN shows the best performance on both CIFAR10 and CIFAR100. Specifically, AE-CNN achieves 4.7\% classification error on CIFAR10, while the best and worst classification error from the peer competitors are 5.4\% and 6.92\%, respectively. In addition, AE-CNN also shows the lower classification error than MetaQNN. AE-CNN shows 0.7\% lower classification error than Large-scale Evolution, and has 32M number of parameters which is also smaller than that of Large-scale Evolution. Furthermore, AE-CNN also consumes much less GPU Days than Large-scale Evolution, NAS and MetaQNN. The comparison shows that the proposed algorithm achieves the best performance among the automatic peer competitors to which the proposed algorithm belongs.

\section{Conclusions and Future Work}
\label{section6}
The goal of this paper is to develop a CNN architecture design algorithm by using GAs, which is capable of designing/finding/learning/evolving the best CNN architecture for the given task with the \textit{\textbf{completely automatic}} way and based on the limited computational resource. This goal has been successfully achieved by the proposed encoding strategy built on the state-of-the-art blocks with a variable-length representation, presenting a crossover operator for the variable-length individuals, and the corresponding mutation operators. Building upon the blocks is able to speed up the CNN architecture design. The variable-length of individuals can adaptively evolve the proper depth of a CNN for tasks with different complexity. The presented crossover operator and the designed mutation operators provide the proposed algorithm with the local search and global search ability, which helps the proposed algorithm to be able to find the best CNN architectures. The proposed algorithm is examined on CIFAR10 and CIFAR100 image classification datasets, against nine state-og-the-art CNNs manually designed, four peer competitors designing CNN architectures with a semi-automatic way and five peer competitors designing CNN architectures with the completely automatic way. The results show that the proposed algorithm outperforms all the state-of-the-art CNNs hand-drafted and all the peer competitors from the automatic category in terms of the classification accuracy. In addition, the proposed algorithm also consumes a much smaller number of GPU Days than the peer competitors in this category. Furthermore, the proposed algorithm shows competitive performance against the semi-automatic peer competitors. The future work will focus on speeding up the fitness evaluation.

\bibliographystyle{aaai}

\end{document}